\pdfoutput=1

\documentclass[11pt]{article}

\usepackage{naacl2021}

\usepackage{times}
\usepackage{latexsym}

\usepackage[T1]{fontenc}

\usepackage[utf8]{inputenc}

\usepackage{microtype}

\usepackage{graphicx}
\usepackage{amsfonts}
\usepackage{amsmath}
\usepackage{booktabs}
\usepackage{multirow}
\usepackage{tabularx, colortbl}

\usepackage{arydshln}

\definecolor{DarkGreen}{RGB}{51,138,60}
\definecolor{DarkRed}{RGB}{200,0,0}
\definecolor{LeftBlue}{RGB}{165,199,232}
\definecolor{RightRed}{RGB}{196,156,153}

\title{Inference Time Style Control for Summarization}

\author{Shuyang Cao \and Lu Wang \\
  Computer Science and Engineering \\
  University of Michigan \\
  Ann Arbor, MI \\
  \texttt{\{caoshuy, wangluxy\}@umich.edu} \\}

\begin{document}
\maketitle

\begin{abstract}

How to generate summaries of different styles without requiring corpora in the target styles, or training separate models? 
We present two novel methods that can be deployed during summary decoding on any pre-trained Transformer-based summarization model.
(1) \textit{Decoder state adjustment} instantly modifies decoder final states with externally trained style scorers, to iteratively refine the output against a target style. 
(2) \textit{Word unit prediction} constrains the word usage to impose strong lexical control during generation.
In experiments of summarizing with simplicity control,  automatic evaluation and human judges both find our models producing outputs in simpler languages while still informative. We also generate news headlines with various ideological leanings, which can be distinguished by humans with a reasonable probability.

\end{abstract}
\section{Introduction}

Generating summaries with different language styles can benefit readers of varying literacy levels~\cite{chandrasekaran-etal-2020-overview} or interests~\cite{jin-etal-2020-hooks}. 
Significant progress has been made in abstractive summarization with large pre-trained Transformers~\cite{NIPS2019_9464,lewis-etal-2020-bart,zhang2019pegasus,raffel2019exploring,song19d}. 
However, style-controlled summarization is much less studied~\cite{chandrasekaran-etal-2020-overview},
and two key challenges have been identified:
(1) \textit{lack of parallel data}, 
and (2) \textit{expensive (re)training}, e.g., separate summarizers must be trained or fine-tuned for a pre-defined set of styles~\cite{zhang-etal-2018-shaped}.  
Both challenges call for inference time methods built upon trained summarization models, to adjust styles flexibly and efficiently.

\begin{figure}[t]
    \fontsize{9}{11}\selectfont
    \setlength{\tabcolsep}{0pt}
    \hspace{-1mm}
    \begin{tabular}{p{0.49\textwidth}}
        \toprule
        \textbf{Daily Mail Article}: $\ldots$ $\pmb{\big [}$ A 16-year-old \textcolor{gray}{who} was born a girl but identifies as a boy has been \textcolor{cyan!10!blue}{\textbf{granted the opportunity}} \textcolor{gray}{to go through male puberty thanks to} hormone treatment. $\pmb{\big ]}$
        $\ldots$ $\pmb{\big [}$ \textcolor{gray}{The transgender boy, who has felt as though he is living in the wrong body since he was a child,} has been given permission by a \textcolor{gray}{Brisbane-based} judge to receive testosterone injections $\pmb{\big ]}$ 
        $\ldots$ \\
        \midrule
        \textbf{(a) Decoder State Adjustment}:  $\pmb{\big [}$ Queensland teen has been granted hormone treatment. The 16-year-old was born a girl but identifies as a boy. $\pmb{\big ]}$
        $\ldots$
        $\pmb{\big [}$ A judge has granted the teen permission to receive testosterone injections. $\pmb{\big ]}$ $\ldots$ \\
        \textbf{(b) Word Unit Prediction}: A 16-year-old who was born a girl has been \textcolor{cyan!10!blue}{\textbf{given the right}} to go through male puberty. The transgender boy has lived in a female body since he was a $\ldots$ \\
        \bottomrule
    \end{tabular}
    \vspace{-1mm}
    \caption{ 
    Sample summaries generated by our style control methods via (a) adjusting decoder states with a simplicity scorer and (b) predicting simple words to use.
    \textcolor{gray}{\bf Gray} texts are produced by BART but removed after decoder state adjustment. Simplified words and their counterparts in the source are highlighted in \textcolor{cyan!10!blue}{\bf blue}. 
    }
    \label{fig:intro}
    \vspace{-2mm}
\end{figure}

To address these challenges, we investigate \textit{just-in-time style control techniques that can be directly applied to any pre-trained sequence-to-sequence (seq2seq) summarization model}. 
We study two methods that leverage external classifiers to favor the generation of words for a given style. 
First, \textbf{decoder state adjustment} is proposed to alter the decoder final states with feedback signaled by style scorers, which are trained to capture global property. 
Second, to offer stronger {\it lexical control}, we introduce \textbf{word unit prediction} that directly constrains the output vocabulary.
Example system outputs are displayed in Fig.~\ref{fig:intro}. 
Notably, our techniques are deployed at \textit{inference time} so that the summary style can be adaptively adjusted during decoding.

We experiment with two tasks: (1) {simplicity control for document summarization} with CNN/Daily Mail, and (2) {headline generation with various ideological stances} on news articles from the SemEval task~\cite{stein:2019i} and a newly curated corpus consisting of multi-perspective stories from AllSides\footnote{\url{www.allsides.com}}. 
In this work, the algorithms are experimented with the BART model~\cite{lewis-etal-2020-bart}, though they also work with other Transformer models. 
Both automatic and human evaluations show that our models produce summaries in simpler languages than competitive baselines, and the informativeness is on par with a vanilla BART. Moreover, headlines generated by our models embody stronger ideological leaning than nontrivial comparisons.\footnote{Our code and data are available at: \url{https://shuyangcao.github.io/projects/inference_style_control}.} 

\section{Related Work}

\noindent \textbf{Summarizing documents into different styles} are mainly studied on news articles, where one appends style codes as extra embeddings to the encoder~\cite{fan-etal-2018-controllable}, or connects separate decoders with a shared encoder~\cite{zhang-etal-2018-shaped}. 
Similar to our work, \citet{jin-etal-2020-hooks} leverage large pre-trained seq2seq models, but they modify model architecture by adding extra style-specific parameters. 
Nonetheless, existing work requires training \textit{new} summarizers for different target styles or modifying the model structure. 
In contrast, our methods only affect decoder states or lexical choices during inference, allowing on-demand style adjustment for summary generation. 

\noindent \textbf{Style-controlled text generation} has received significant research attentions, especially where parallel data is scant~\cite{lample2018multipleattribute,shang-etal-2019-semi,He2020A}.
Typical solutions involve disentangling style representation from content representation, and are often built upon autoencoders~\cite{pmlr-v70-hu17e} with adversarial training objectives~\cite{NIPS2018_7959}. The target style is then plugged in during generation. 
Recently, \citet{Dathathri2020Plug} propose plug and play language models (\textsc{PPLM}s) to alter the generation style by modifying all key-value pairs in the Transformer, which requires heavy computation during inference. \citet{krause2020gedi} then employ a generative discriminator (GeDi) to improve efficiency. Our methods are more efficient since we only modify the decoder final states or curtail the vocabulary.

\section{Inference Time Style Control}

\begin{figure}[t]
    \centering
    \includegraphics[width=0.42\textwidth]{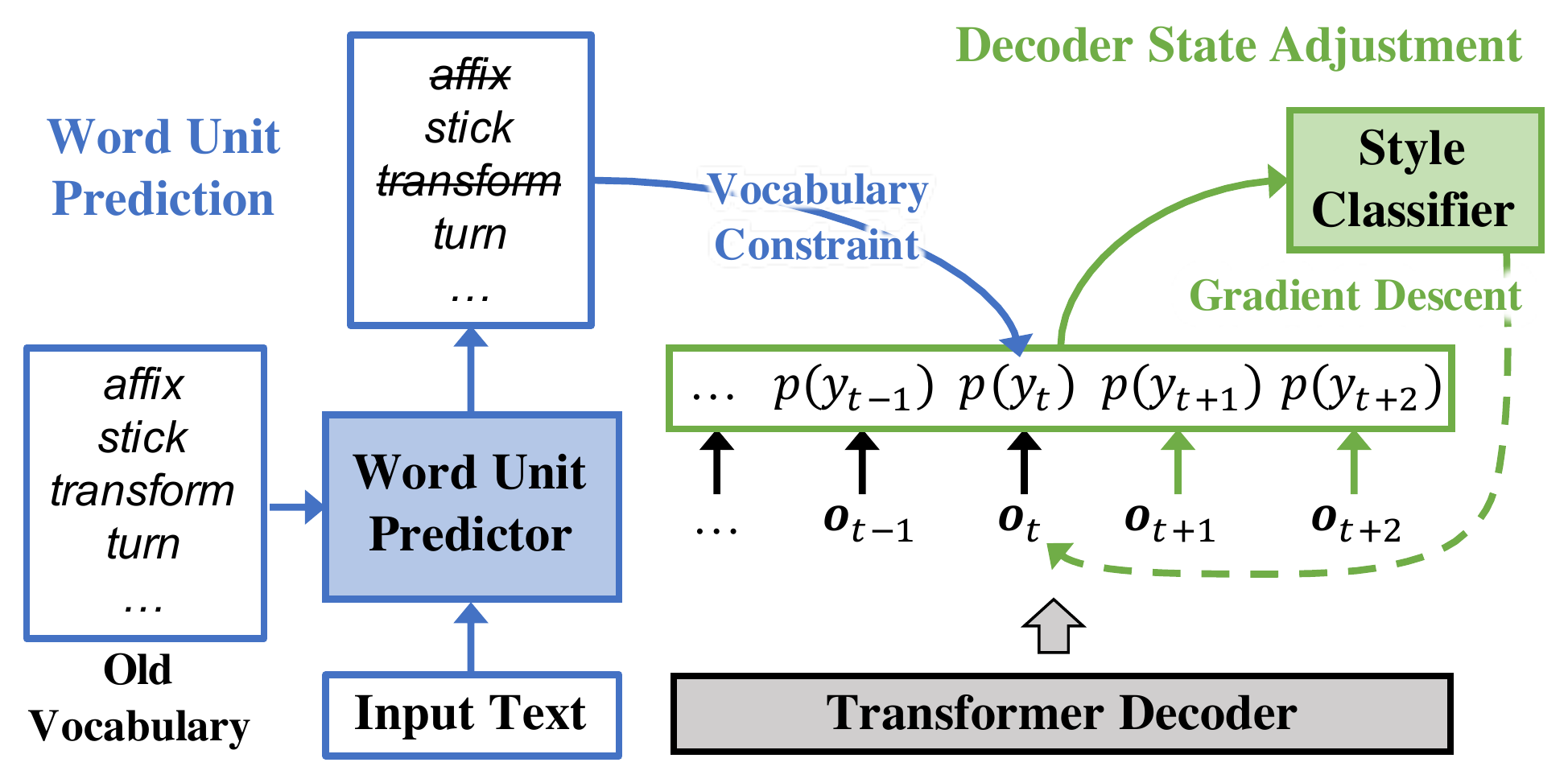}
    \vspace{-1mm}
    \caption{
    Just-in-time style control: 
    (1) Decoder state adjustment takes in a style score and iteratively updates $\mathbf{o}_{t}$; 
    (2) Word unit prediction controls the vocabulary. 
    }
    \label{fig:style_control}
    \vspace{-3mm}
\end{figure}

\subsection{Global Characteristic Control via Decoder State Adjustment} 
\label{subsec:global_control}

Given a style classifier $q(z \vert \cdot)$ that measures to which extent does the current generated summary resemble the style $z$, 
we use its estimate to adjust the final decoder layer's state $\mathbf{o}_t$ at step $t$ with gradient descent, as illustrated in Fig.~\ref{fig:style_control}. 
The output token is produced as $p(y_t \vert y_{1:t-1}, \mathbf{x}) = \text{softmax} ( \mathbf{W}_{e} \mathbf{o}_t )$, $\mathbf{W}_{e}$ is the embedding matrix. 

Concretely, to generate the $t$-th token, a style score of $q(z \vert \mathbf{y}_{1:t+2})$ is first computed. 
In addition to what have been generated up to step $t-1$, we also sample $y_t$ and two future tokens for style estimation. 
The decoder state is updated as follows:

\vspace{-1mm}
{
    \fontsize{10}{11}\selectfont
    \begin{equation}
        \mathbf{o}_t \gets \mathbf{o}_t - \lambda \nabla_{\mathbf{o}_t} \big[- q(z \vert \mathbf{y}_{1:t+2}) \big]
    \end{equation}
}%
where $\lambda$ is the step size. Gradient descent is run for $10$ iterations for document summarization and $30$ iterations for headline generation.

Below, we define one discriminative and one generative style classifier, to illustrate the method.

\smallskip
\noindent \textbf{Discriminative Style Scorer.} 
We feed the tokens into a RoBERTa encoder~\cite{liu2019roberta} and use the contextualized representation of the \texttt{BOS} token, i.e., $\mathbf{h}_0$, to predict the style score as $p_{sty} (z \vert \cdot) = \mathrm{softmax} (\mathbf{W}_{s}\mathbf{h}_0)$, where $\mathbf{W}_{\ast}$ are learnable parameters in this paper. At step $t$ of summary decoding, the style score is estimated as:

\vspace{-1mm}
{
    \fontsize{10}{11}\selectfont
    \begin{equation}
        q(z \vert \mathbf{y}_{1:t+2}) = \log p_{sty}(z \vert \mathbf{y}_{1:t+2})
    \end{equation}
}%
For the discriminative style scorer, the step size $\lambda$ is set to $1.0$.

\smallskip
\noindent \textbf{Generative Language Model Scorer.}
We build a class-conditional language model (CC-LM) from texts prepended with special style-indicating tokens.
Concretely, the CC-LM yields probabilities $p_{LM} (y_{t^\prime} \vert \mathbf{y}_{1:t^\prime-1}, z)$ ($p_{LM}(y_{t^\prime}, z)$ for short), conditional on the previously generated tokens $\mathbf{y}_{1:t^\prime-1}$ and the style $z$.
As the summarizer's output probability $p(y_{t^\prime})$ should be close to the language model's estimate, the style score is defined as: 

\vspace{-5mm}
{
    \fontsize{10}{11}\selectfont
    \begin{equation}
        q(z \vert \mathbf{y}_{1:t+2}) = \frac{1}{t+2} \sum^{t+2}_{t^\prime=1} p_{LM}(y_{t^\prime}, z) \log p(y_{t^\prime})
    \end{equation}
}%
Here we use a step size $\lambda$ of $0.1$.

\subsection{Lexical Control via Word Unit Prediction} 
\label{subsec:lexical_control}

Lexical control is another tool for managing summary style, as word choice provides a strong signal of language style.
Given an input document, our goal is to predict a set of word units (e.g., the subwords used in BART pre-training) that can be used for summary generation. For instance, if the input contains ``affix'', we will predict ``stick'' to be used, while excluding the original word ``affix''. A similar idea has been used to expedite sequence generation~\cite{hashimoto-tsuruoka-2019-accelerated}, though our goal here is to calculate the possibilities of different lexical choices.

Concretely, after encoding the input $\mathbf{x}$ by RoBERTa, we take the average of all tokens' contextual representations, and pass it through a residual block~\cite{he2016deep} to get its final representation $\mathbf{\tilde{R}}$. We then compute a probability vector for all word units in the vocabulary as $\mathbf{p}^{r} = \text{sigmoid} (\mathbf{W}_r \mathbf{\tilde{R}})$. 
The top $v$ word units with the highest probabilities 
are selected and combined with entity names from the input to form the new vocabulary, from which the summary is generated.
We use $v=1000$ in all experiments.

\smallskip
\noindent \textbf{Dynamic Prediction.} 
We also experiment with a dynamic version, where the word unit predictor further considers what have been generated up to a given step. In this way, the new vocabulary is updated every $m$ steps ($m=5$ for document summarization, and $m=3$ for headline generation).

\section{Simplicity-controlled Document Summarization}

For experiments, we use BART fine-tuned on the CNN/DailyMail (CNN/DM)~\cite{NIPS2015_5945}, by following \citet{lewis-etal-2020-bart} for data preprocessing and splitting. The numbers of data in train, validation and test splits are $287{,}188$, $13{,}367$ and $11{,}490$, respectively.

We use paragraph pairs from normal and simple English Wikipedia articles in~\citet{hua-wang-2019-sentence} for {\it simplicity style scorer} and {\it class-conditional language model} training. We split the pairs into $86{,}467$, $10{,}778$, and $10{,}788$ for training, validation and testing, respectively. On the test set, our simplicity style scorer achieves an F1 score of $89.7$ and our class-conditional language model achieves a perplexity of $30.35$.

To learn the {\it word unit predictor}, for each paragraph pair, the predictor reads in the normal version and is trained to predict the word units used in the simple version.
For the {\it dynamic version}, it predicts which word units are used to generate the rest of the text, after every $5$ steps. Recalls for the two predictors on the test set are $81.5$ and $80.0$.

For comparison, we consider \textbf{\textsc{Reranking}} beams based on our style score at the last step. 
We also use a label-controlled (\textbf{\textsc{LblCtrl}}) baseline as described in \citet{niu-bansal-2018-polite}, where summaries in the training data are labeled as simple or normal by our scorer. 
We further compare with \textbf{\textsc{GeDi}} and two pipeline models: a style transfer model~\cite{pmlr-v70-hu17e} applied on the output of BART (\textbf{\textsc{CtrlGen}}) and a normal-to-simple translation model fine-tuned from BART (\textbf{\textsc{Trans}}), both trained on Wikipedia.
Finally, we consider \textbf{\textsc{LightLS}}~\cite{glavavs-vstajner:2015:ACL-IJCNLP}, a rule-based lexical simplification model.

\begin{table}[t]
\small
    \centering
    \setlength{\tabcolsep}{1.3pt}
    \begin{tabular}{lccccc}
    \toprule
        \multirow{2}{*}{\textbf{Model}} & \multicolumn{3}{c}{\textbf{Style}} & \textbf{Flu.} & \textbf{Cont.} \\
        \cmidrule(lr){2-4}
        \cmidrule(lr){5-5}
        \cmidrule(lr){6-6}
        & \textbf{Simp.$\uparrow$} & \textbf{\%Simp.$\uparrow$} & \textbf{Rd.$\downarrow$} & \textbf{PPL$\downarrow$} & \textbf{BERT$\uparrow$} \\
        \midrule
        \textsc{BART} & 56.93 & 62.70 & 8.06 & 34.05 & 88.62 \\
        \midrule
        \textsc{Reranking} & 71.33 & 62.68 & 8.04 & 36.17 & 88.62 \\
        \textsc{LblCtrl} & 56.21 & 62.71 & 8.07 & 28.85 & 88.57 \\
        \textsc{CtrlGen} & 81.56 & 64.78 & 7.79 & 70.36 & 88.01 \\
        \textsc{Trans} & 59.78 & 63.03 & 7.99 & 33.17 & 88.46 \\
        \textsc{GeDi} & 71.33 & 62.57 & 7.88 & 33.48 & \textbf{88.79} \\
        \textsc{LightLS} & 69.02 & 64.92 & 7.72 & 76.37 & 86.98 \\
        \hdashline
        \multicolumn{6}{l}{\rule{0pt}{2ex}\bf Ours w/ Decoder State Adjustment}\\
        \textsc{Simp. Scorer} & 86.67 & 62.94 & 7.77 & 34.20 & 88.71 \\
        \textsc{Simp. CC-LM} & 75.04 & 64.27 & 7.69 & 30.49 & 88.73 \\
        \multicolumn{6}{l}{\bf Ours w/ Word Unit Prediction}\\
        \textsc{WordU} & \textbf{95.85} & 67.23 & \textbf{7.19} & \textbf{27.40} & 87.76 \\
        \textsc{Dynamic WordU} & 93.87 & \textbf{67.37} & 7.23 & 28.42 & 87.91 \\
        \bottomrule
    \end{tabular}
    \vspace{-1mm}
    \caption{
    Automatic evaluation on summarization with simplicity, with simplicity level by our scorer (Simp., probability multiplied by 100),  \% of words in the Dale-Chall simple word list (\%Simp.), Dale-Chall readability (Rd.), fluency by perplexity (PPL), and content metric by BERTScore (BERT). Our models are significantly better than the comparisons ($p < 0.005$) on simplicity and readability, except for \textsc{CtrlGen} and \textsc{LightLS}.
    }
    \label{tab:doc_summ_simplicity_auto_eval}
    \vspace{-2mm}
\end{table}

\smallskip
\noindent \textbf{Automatic Evaluation.}
Table~\ref{tab:doc_summ_simplicity_auto_eval} shows that \textit{our models' outputs have significantly better simplicity and readability while preserving fluency and a comparable amount of salient content}. 
Key metrics include simplicity level estimated by our scorer and Dale-Chall readability~\cite{chall1995readability}. 
We use GPT-2 perplexity~\cite{radford2019language} to measure fluency, and BERTScore~\cite{Zhang*2020BERTScore:} for content preservation. 
Our inference time style control modules can adaptively change the output style, and thus outperform reranking at the end of generation or using pipeline models. Moreover, by iteratively adjusting the decoder states, our methods deliver stronger style control than \textsc{GeDi}, which only adjusts the probability once per step.

When comparing among our models, we find that \textit{word unit prediction is more effective at lexical simplification than updating decoder states}, as demonstrated by the higher usage of simple words according to the Dale-Chall list. We believe that strong lexical control is achieved by directly pruning output vocabulary, whilst decoder state adjustment is more poised to capture global property, e.g., sentence compression as shown in Fig.~\ref{fig:intro}. Moreover, we compute the edit distance between our style-controlled system outputs and the summaries produced by the fine-tuned BART. We find that adjusting decoder states with style scorer and language model yields an edit distance of $45.7$ and $47.4$, compared to larger distances of $56.7$ and $54.3$ given by word unit prediction and with additional dynamic prediction.

\smallskip
\noindent \textbf{Human Evaluation.}
We recruit three fluent English speakers to evaluate system summaries for \textbf{informativeness}---whether the summary covers important information from the input, and \textbf{fluency}---whether the summary is grammatical, 
on a scale of 1 (worst) to 5 (best). 
They then rank the summaries by \textbf{simplicity} level (ties are allowed). 
50 samples are randomly selected for evaluation, and system summaries are shuffled. 
As seen in Table~\ref{tab:simplicity_human_eval}, \textit{summaries by our models are considered simpler} than outputs of BART and \textsc{GeDi}, with better or comparable informativeness.

\begin{table}[t]
    \small
    \centering
    \setlength{\tabcolsep}{3pt}
    \begin{tabular}{lcccc}
    \toprule
        \textbf{Model} & \textbf{Inf.$\uparrow$} & \textbf{Flu.$\uparrow$} & \textbf{Simp.R.$\downarrow$} & \textbf{Top 1$\uparrow$} \\
        \midrule
        \textsc{BART} & 4.45 & \textbf{4.90} & 2.19 & 19.0\% \\
        \textsc{GeDi} & 4.48 & 4.83 & 2.00 & 23.8\% \\
        \hdashline
        \rule{-2pt}{2ex}
        \textsc{Simp. Scorer} & \textbf{4.53} & 4.83 & 1.66{$^\ast$} & 48.4\% \\
        \textsc{Dynamic WordU}  & 4.36 & 4.84 & \textbf{1.65}{$^\ast$} & \textbf{57.9\%} \\
        \bottomrule
    \end{tabular}
    \vspace{-1mm}
    \caption{
    Human evaluation on informativeness (Inf.), fluency (Flu.), simplicity ranking (Simp.R.), and percentage of summaries ranked as simplest (Top 1). 
    Krippendorff's $\alpha$: 0.38, 0.22, and 0.16 (first three metrics). 
    $\ast$: significantly better than comparisons ($p < 0.005$). 
    }
    \label{tab:simplicity_human_eval}
    \vspace{-3mm}
\end{table}

\section{Ideology-controlled Headline Generation}

To generate news headlines of various ideological leanings, we use the \textbf{SemEval} Hyperpartisan News Detection dataset~\cite{stein:2019i}, where each article is labeled with a stance: \textit{left}, \textit{leaning left}, \textit{neutral}, \textit{leaning right}, or \textit{right}. Here, we combine left and leaning-left articles into one bucket, and similarly for right and leaning-right articles.
We use the lead paragraph as the input, and the headline as the target generation. The data is processed following~\citet{rush-etal-2015-neural}, and split into $346{,}985$ for training, $30{,}000$ each for validation and testing. Details of the ideology distribution for SemEval are in Appendix~\ref{append:semeval_allsides}.

We fine-tune BART and train ideology classifiers on the SemEval training set.
First, two binary {\it style scorers} are trained on headlines of left and right stances, with F1 scores of $76.1$ and $78.0$, respectively.
One {\it class-conditional language model} is trained on headlines with a stance token (left or right) prepended, achieving a perplexity of $54.7$.
To learn the {\it word unit predictor} for the left (and similarly for the right), we use samples that are labeled as left-leaning, treat the lead paragraph as the input, and then predict the word units used in the headline. Recalls for our predictors range from $77.8$ to $83.5$.

\begin{table}[t]
\small
    \centering
    \setlength{\tabcolsep}{4pt}
    \begin{tabular}{lccccc}
    \toprule
        \multirow{2}{*}{\textbf{Model}} & \multicolumn{2}{c}{\textbf{Left}} & \multicolumn{2}{c}{\textbf{Right}} \\
        \cmidrule(lr){2-3}
        \cmidrule(lr){4-5}
        & \textbf{Ideol.} & \textbf{BERT} & \textbf{Ideol.} & \textbf{BERT}  \\
        \midrule
        \textsc{BART}  & 18.63 & 91.03 & 19.04 & 91.03 \\
        \midrule
        \textsc{Reranking} & 30.80 & 90.68 & 30.11 & 90.66 \\
        \textsc{LblCtrl} & 20.59 & \textbf{90.97} & 20.89 & \textbf{91.02} \\
        \textsc{GeDi} & 12.64 & 84.84 & 3.61 & 84.84 \\
        \hdashline
        \multicolumn{6}{l}{\rule{0pt}{2ex}\bf Ours w/ Decoder State Adjustment}\\
        \textsc{Ideol. Scorer} & \textbf{31.15} & 90.08 & \textbf{30.54} & 90.17 \\
        \textsc{Ideol. CC-LM} & 23.74 & 89.65 & 20.79 & 89.65 \\
        \multicolumn{6}{l}{\bf Ours w/ Word Unit Prediction}\\
        \textsc{WordU} & 21.30 & 89.64 & 20.42 & 90.13 \\
        \textsc{Dynamic WordU} & 21.53 & 89.49 & 20.09 & 90.19 \\
        \bottomrule
    \end{tabular}
    \vspace{-1mm}
    \caption{
    Ideological headline generation results.
    Using ideology scorer to update decoder states yields the highest ideology scores (multiplied by 100).  
    }
    \label{tab:ideology_auto_eval}
    \vspace{-1mm}
\end{table}

\begin{figure}[t]
    \centering
    \includegraphics[width=0.45\textwidth]{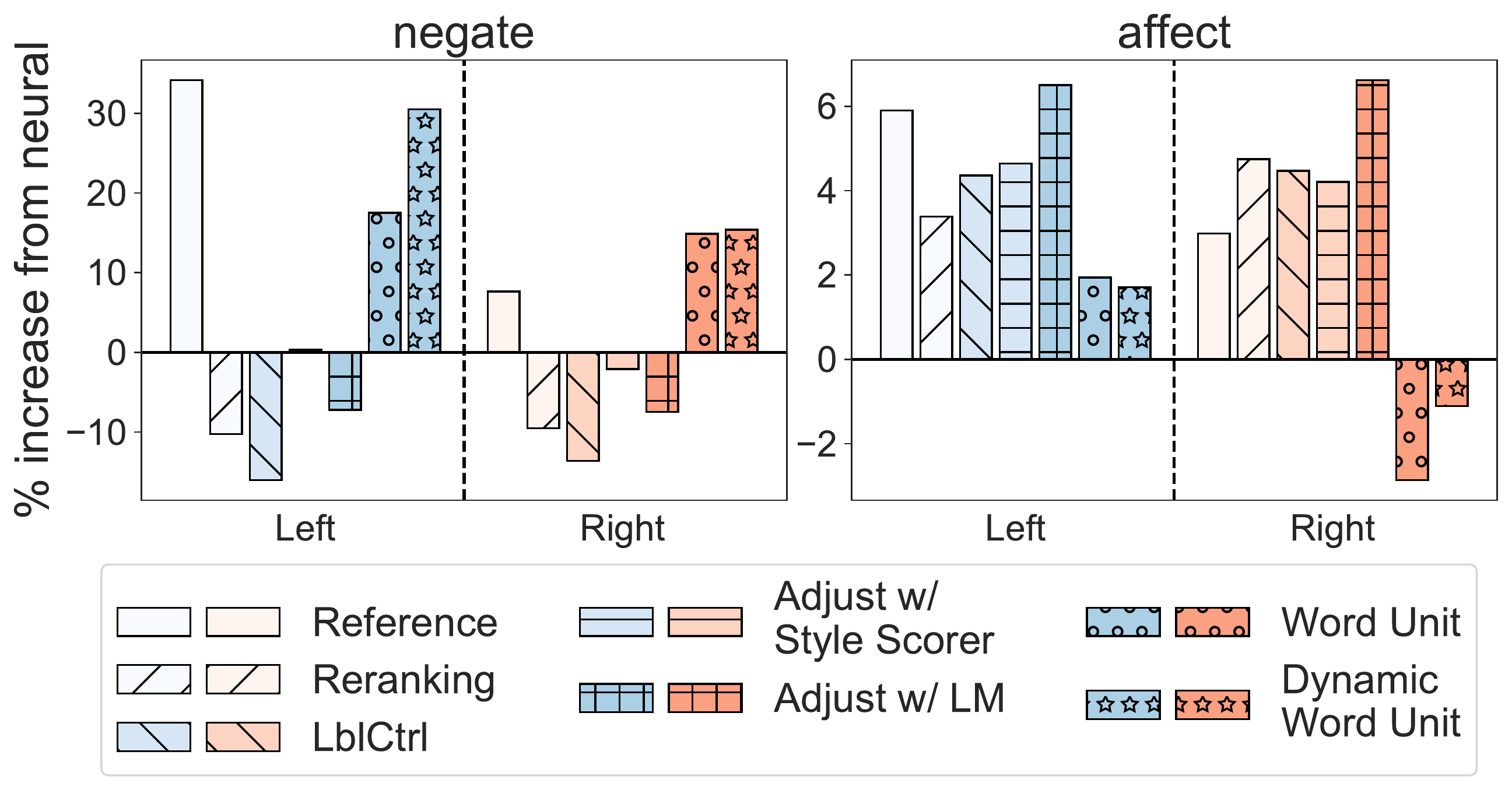}
    \vspace{-1mm}
    \caption{
    LIWC word usage changes of ``negate'' and ``affect'', compared to neutral headlines. In each subfigure, left and right panels correspond to left and right leaning stances. 
    }
    \label{fig:liwc}
    \vspace{-2mm}
\end{figure}

\smallskip
\noindent \textbf{Automatic Evaluation with SemEval.}
Table~\ref{tab:ideology_auto_eval} shows that \textit{our decoder state adjustment model with the ideology scorer obtains the highest ideology scores}, due to its effectiveness at capturing the global context---stance is often signaled by the joint selection of entities and sentiments. 

One might be interested in \textit{which words are favored for ideology-controlled generation}. To that end, we analyze the change of word usages with Linguistic Inquiry and Word Count (LIWC)~\cite{pennebaker2015development}. 
In Fig.~\ref{fig:liwc}, it can be seen that word unit prediction-based models generate more ``negations'', consistent with trends observed in human-written headlines. 
Meanwhile, models with decoder state adjustment and the baselines all use more ``affect'' words in both stances, indicating that they consider it easier to use explicit sentiments to demonstrate the stances.

\begin{table}[t]
    \small
    \setlength{\tabcolsep}{2pt}
    \centering
    \begin{tabular}{llllc}
    \toprule
        \textbf{Model} & \textbf{Rel.} & \textbf{Edit} & \textbf{Hmn} & \textbf{Hmn Acc.}\\
        \midrule
        Human & 4.01 & 12.24 & 60.8\% & 73.3\% \\
        \midrule
        \textsc{Reranking} & \textbf{4.71} & 3.90 & 24.5\% & 52.5\% \\
        \textsc{LblCtrl} & 4.70 & 2.30 & 11.6\% & \textbf{71.4\%} \\
        \hdashline
        \rule{-2pt}{2ex}
        \textsc{Ideol. Scorer} (ours) & 4.47 & \textbf{8.86}{$^\ast$} & \textbf{42.5\%}{$^\ast$} & 53.9\% \\
        \textsc{Dynamic WordU} (ours) & 4.66 & 4.20 & 25.8\% & 51.6\% \\
        \bottomrule
    \end{tabular}
    \vspace{-1mm}
    \caption{
    Human evaluation of ideology-controlled headline generation with relevance (Rel.), edit distance (Edit) between left and right headlines, \% of samples perceived as having different stances (Hmn), and (among them) accuracy of identified stances (Hmn Acc.). 
    Krippendorff's $\alpha$ of relevance: 0.48.
    $\ast$: significantly better than other models ($p < 0.005$).
    }
    \label{tab:ideology_human_eval}
    \vspace{-2mm}
\end{table}

\smallskip
\noindent \textbf{Human Evaluation with AllSides.} 
Given the low ideology scores in Table \ref{tab:ideology_auto_eval}, we further study \textit{if human can distinguish the stances in human-written and system generated headlines}. 
News clusters from \textbf{AllSides} are used, where each cluster focuses on one story, with multiple paragraph-headline pairs from publishers of \textit{left}, \textit{neutral}, and \textit{right} ideological leanings. We use the lead paragraph as the input, and collect $2{,}985$ clusters with samples written in all three stances. More details of the collection are in Appendix~\ref{append:semeval_allsides}.
We test and report results by using lead paragraphs from \textit{neural} articles as the input to construct headlines of \textit{left} and \textit{right} ideological stances. 
 
We randomly pick 80 samples and include, for each sample, two headlines of different stances generated by each system. 
Raters first score the \textbf{relevance} of the generated headlines to the neutral paragraph's headline, on a scale of $1$ to $5$. They then read each pair of headlines to decide whether they are written in different stances, and if so, to label them. 
Table~\ref{tab:ideology_human_eval} highlights the intrinsic difficulty of capturing ideological language usage: Even reference headlines are only distinguishable in $60.8\%$ of the cases, among which the stance identification accuracy is $73.3\%$.
In comparison, $42.5\%$ of the output pairs by the decoder state adjustment model can be distinguished, significantly higher than those of the baselines ($24.5\%$ and $11.6\%$). 
Sample outputs by our models are shown in Table~\ref{tab:sample_ideol}, with more outputs included in Appendix~\ref{append:sample_output}.

\begin{table}[t]
    \small
    \setlength{\tabcolsep}{0pt}
    \begin{tabular}{p{0.49\textwidth}}
    \toprule
    \textbf{Paragraph}: The Obama administration on Thursday rolled out new efforts aimed at curtailing gun violence $\ldots$ \\
    \midrule
        \textbf{\textsc{Reference}} \\
        \rowcolor{LeftBlue!60}
        $[$L$]$: obama offers new executive actions on gun control \\
        \rowcolor{red!15}
        $[$R$]$: administration announces new gun control measures, targets military surplus imports \\
        \midrule
        \textbf{\textsc{Ideol.Scorer}} \\
        \rowcolor{LeftBlue!60} $[$L$]$: u.s. moves to \textbf{curb gun violence} with new rules \\
        \rowcolor{red!15} $[$R$]$: obama admin to \textbf{tighten gun control laws} \\
        \midrule
        \textbf{\textsc{Dynamic WordU}} \\
        \rowcolor{LeftBlue!60} $[$L$]$: obama unveils new steps to \textbf{curb gun violence} \\
        \rowcolor{red!15} $[$R$]$: obama administration unveils new \textbf{gun control} measures \\
        \bottomrule
    \end{tabular}
    \vspace{-1mm}
    \caption{
    Sample generated headlines with left (shaded in \textcolor{LeftBlue!99}{\bf blue}) and right (\textcolor{red!60}{\bf red}) stances. Phrases that are typically used by a stance are in \textbf{bold}.
    }
    \label{tab:sample_ideol}
    \vspace{-2mm}
\end{table}

\section{Conclusion}

We present two just-in-time style control methods, which can be used in any Transformer-based summarization models. The decoder state adjustment technique modifies decoder final states based on externally trained style scorers. To gain stronger lexical control, word unit prediction directly narrows the vocabulary for generation. 
Human judges rate our system summaries to be simpler with better readability. We are also able to generate headlines with different ideological leanings.
\section*{Acknowledgements}
This research is supported in part by National Science Foundation through Grant IIS-1813341, and by the Office of the Director of National Intelligence (ODNI), Intelligence Advanced Research Projects Activity (IARPA), via contract \# FA8650-17-C-9116. The views and conclusions contained herein are those of the authors and should not be interpreted as necessarily representing the official policies, either expressed or implied, of ODNI, IARPA, or the U.S. Government. The U.S. Government is authorized to reproduce and distribute reprints for governmental purposes notwithstanding any copyright annotation therein. 
We thank all the anonymous reviewers for their constructive suggestions. 

\newpage

\bibliography{custom}

\begin{thebibliography}{31}
\expandafter\ifx\csname natexlab\endcsname\relax\def\natexlab#1{#1}\fi

\bibitem[{Chall and Dale(1995)}]{chall1995readability}
J.S. Chall and E.~Dale. 1995.
\newblock \href {https://books.google.com/books?id=2nbuAAAAMAAJ}
  {\emph{Readability revisited: the new Dale-Chall readability formula}}.
\newblock Brookline Books.

\bibitem[{Chandrasekaran et~al.(2020)Chandrasekaran, Feigenblat, Freitag,
  Ghosal, Hovy, Mayr, Shmueli-Scheuer, and
  de~Waard}]{chandrasekaran-etal-2020-overview}
Muthu~Kumar Chandrasekaran, Guy Feigenblat, Dayne Freitag, Tirthankar Ghosal,
  Eduard Hovy, Philipp Mayr, Michal Shmueli-Scheuer, and Anita de~Waard. 2020.
\newblock \href {https://www.aclweb.org/anthology/2020.sdp-1.1} {Overview of
  the first workshop on scholarly document processing ({SDP})}.
\newblock In \emph{Proceedings of the First Workshop on Scholarly Document
  Processing}, pages 1--6, Online. Association for Computational Linguistics.

\bibitem[{Dathathri et~al.(2020)Dathathri, Madotto, Lan, Hung, Frank, Molino,
  Yosinski, and Liu}]{Dathathri2020Plug}
Sumanth Dathathri, Andrea Madotto, Janice Lan, Jane Hung, Eric Frank, Piero
  Molino, Jason Yosinski, and Rosanne Liu. 2020.
\newblock \href {https://openreview.net/forum?id=H1edEyBKDS} {Plug and play
  language models: A simple approach to controlled text generation}.
\newblock In \emph{International Conference on Learning Representations}.

\bibitem[{Dong et~al.(2019)Dong, Yang, Wang, Wei, Liu, Wang, Gao, Zhou, and
  Hon}]{NIPS2019_9464}
Li~Dong, Nan Yang, Wenhui Wang, Furu Wei, Xiaodong Liu, Yu~Wang, Jianfeng Gao,
  Ming Zhou, and Hsiao-Wuen Hon. 2019.
\newblock \href
  {http://papers.nips.cc/paper/9464-unified-language-model-pre-training-for-natural-language-understanding-and-generation.pdf}
  {Unified language model pre-training for natural language understanding and
  generation}.
\newblock In H.~Wallach, H.~Larochelle, A.~Beygelzimer, F.~d'~Alch\'{e}-Buc,
  E.~Fox, and R.~Garnett, editors, \emph{Advances in Neural Information
  Processing Systems 32}, pages 13063--13075. Curran Associates, Inc.

\bibitem[{Fan et~al.(2018)Fan, Grangier, and Auli}]{fan-etal-2018-controllable}
Angela Fan, David Grangier, and Michael Auli. 2018.
\newblock \href {https://doi.org/10.18653/v1/W18-2706} {Controllable
  abstractive summarization}.
\newblock In \emph{Proceedings of the 2nd Workshop on Neural Machine
  Translation and Generation}, pages 45--54, Melbourne, Australia. Association
  for Computational Linguistics.

\bibitem[{Glava\v{s} and \v{S}tajner(2015)}]{glavavs-vstajner:2015:ACL-IJCNLP}
Goran Glava\v{s} and Sanja \v{S}tajner. 2015.
\newblock \href {http://www.aclweb.org/anthology/P15-2011} {Simplifying lexical
  simplification: Do we need simplified corpora?}
\newblock In \emph{Proceedings of the 53rd Annual Meeting of the Association
  for Computational Linguistics and the 7th International Joint Conference on
  Natural Language Processing (Volume 2: Short Papers)}, pages 63--68, Beijing,
  China. Association for Computational Linguistics.

\bibitem[{Hashimoto and Tsuruoka(2019)}]{hashimoto-tsuruoka-2019-accelerated}
Kazuma Hashimoto and Yoshimasa Tsuruoka. 2019.
\newblock \href {https://doi.org/10.18653/v1/N19-1315} {Accelerated
  reinforcement learning for sentence generation by vocabulary prediction}.
\newblock In \emph{Proceedings of the 2019 Conference of the North {A}merican
  Chapter of the Association for Computational Linguistics: Human Language
  Technologies, Volume 1 (Long and Short Papers)}, pages 3115--3125,
  Minneapolis, Minnesota. Association for Computational Linguistics.

\bibitem[{He et~al.(2020)He, Wang, Neubig, and Berg-Kirkpatrick}]{He2020A}
Junxian He, Xinyi Wang, Graham Neubig, and Taylor Berg-Kirkpatrick. 2020.
\newblock \href {https://openreview.net/forum?id=HJlA0C4tPS} {A probabilistic
  formulation of unsupervised text style transfer}.
\newblock In \emph{International Conference on Learning Representations}.

\bibitem[{He et~al.(2016)He, Zhang, Ren, and Sun}]{he2016deep}
Kaiming He, Xiangyu Zhang, Shaoqing Ren, and Jian Sun. 2016.
\newblock Deep residual learning for image recognition.
\newblock In \emph{Proceedings of the IEEE conference on computer vision and
  pattern recognition}, pages 770--778.

\bibitem[{Hermann et~al.(2015)Hermann, Kocisky, Grefenstette, Espeholt, Kay,
  Suleyman, and Blunsom}]{NIPS2015_5945}
Karl~Moritz Hermann, Tomas Kocisky, Edward Grefenstette, Lasse Espeholt, Will
  Kay, Mustafa Suleyman, and Phil Blunsom. 2015.
\newblock \href
  {http://papers.nips.cc/paper/5945-teaching-machines-to-read-and-comprehend.pdf}
  {Teaching machines to read and comprehend}.
\newblock In C.~Cortes, N.~D. Lawrence, D.~D. Lee, M.~Sugiyama, and R.~Garnett,
  editors, \emph{Advances in Neural Information Processing Systems 28}, pages
  1693--1701. Curran Associates, Inc.

\bibitem[{Hu et~al.(2017)Hu, Yang, Liang, Salakhutdinov, and
  Xing}]{pmlr-v70-hu17e}
Zhiting Hu, Zichao Yang, Xiaodan Liang, Ruslan Salakhutdinov, and Eric~P. Xing.
  2017.
\newblock \href {http://proceedings.mlr.press/v70/hu17e.html} {Toward
  controlled generation of text}.
\newblock In \emph{Proceedings of the 34th International Conference on Machine
  Learning}, volume~70 of \emph{Proceedings of Machine Learning Research},
  pages 1587--1596, International Convention Centre, Sydney, Australia. PMLR.

\bibitem[{Hua and Wang(2019)}]{hua-wang-2019-sentence}
Xinyu Hua and Lu~Wang. 2019.
\newblock \href {https://doi.org/10.18653/v1/D19-1055} {Sentence-level content
  planning and style specification for neural text generation}.
\newblock In \emph{Proceedings of the 2019 Conference on Empirical Methods in
  Natural Language Processing and the 9th International Joint Conference on
  Natural Language Processing (EMNLP-IJCNLP)}, pages 591--602, Hong Kong,
  China. Association for Computational Linguistics.

\bibitem[{Jin et~al.(2020)Jin, Jin, Zhou, Orii, and
  Szolovits}]{jin-etal-2020-hooks}
Di~Jin, Zhijing Jin, Joey~Tianyi Zhou, Lisa Orii, and Peter Szolovits. 2020.
\newblock \href {https://doi.org/10.18653/v1/2020.acl-main.456} {Hooks in the
  headline: Learning to generate headlines with controlled styles}.
\newblock In \emph{Proceedings of the 58th Annual Meeting of the Association
  for Computational Linguistics}, pages 5082--5093, Online. Association for
  Computational Linguistics.

\bibitem[{Kiesel et~al.(2019)Kiesel, Mestre, Shukla, Vincent, Adineh, Corney,
  Stein, and Potthast}]{stein:2019i}
Johannes Kiesel, Maria Mestre, Rishabh Shukla, Emmanuel Vincent, Payam Adineh,
  David Corney, Benno Stein, and Martin Potthast. 2019.
\newblock {SemEval-2019 Task 4: Hyperpartisan News Detection}.
\newblock In \emph{12th International Workshop on Semantic Evaluation (SemEval
  2019)}. Association for Computational Linguistics.

\bibitem[{Krause et~al.(2020)Krause, Gotmare, McCann, Keskar, Joty, Socher, and
  Rajani}]{krause2020gedi}
Ben Krause, Akhilesh~Deepak Gotmare, Bryan McCann, Nitish~Shirish Keskar,
  Shafiq Joty, Richard Socher, and Nazneen~Fatema Rajani. 2020.
\newblock \href {http://arxiv.org/abs/2009.06367} {Gedi: Generative
  discriminator guided sequence generation}.

\bibitem[{Lample et~al.(2019)Lample, Subramanian, Smith, Denoyer, Ranzato, and
  Boureau}]{lample2018multipleattribute}
Guillaume Lample, Sandeep Subramanian, Eric Smith, Ludovic Denoyer,
  Marc'Aurelio Ranzato, and Y-Lan Boureau. 2019.
\newblock \href {https://openreview.net/forum?id=H1g2NhC5KQ}
  {Multiple-attribute text rewriting}.
\newblock In \emph{International Conference on Learning Representations}.

\bibitem[{Lewis et~al.(2020)Lewis, Liu, Goyal, Ghazvininejad, Mohamed, Levy,
  Stoyanov, and Zettlemoyer}]{lewis-etal-2020-bart}
Mike Lewis, Yinhan Liu, Naman Goyal, Marjan Ghazvininejad, Abdelrahman Mohamed,
  Omer Levy, Veselin Stoyanov, and Luke Zettlemoyer. 2020.
\newblock \href {https://doi.org/10.18653/v1/2020.acl-main.703} {{BART}:
  Denoising sequence-to-sequence pre-training for natural language generation,
  translation, and comprehension}.
\newblock In \emph{Proceedings of the 58th Annual Meeting of the Association
  for Computational Linguistics}, pages 7871--7880, Online. Association for
  Computational Linguistics.

\bibitem[{Liu et~al.(2019)Liu, Ott, Goyal, Du, Joshi, Chen, Levy, Lewis,
  Zettlemoyer, and Stoyanov}]{liu2019roberta}
Yinhan Liu, Myle Ott, Naman Goyal, Jingfei Du, Mandar Joshi, Danqi Chen, Omer
  Levy, Mike Lewis, Luke Zettlemoyer, and Veselin Stoyanov. 2019.
\newblock Roberta: A robustly optimized bert pretraining approach.
\newblock \emph{arXiv preprint arXiv:1907.11692}.

\bibitem[{Napoles et~al.(2012)Napoles, Gormley, and
  Van~Durme}]{napoles-etal-2012-annotated}
Courtney Napoles, Matthew Gormley, and Benjamin Van~Durme. 2012.
\newblock \href {https://www.aclweb.org/anthology/W12-3018} {Annotated
  {G}igaword}.
\newblock In \emph{Proceedings of the Joint Workshop on Automatic Knowledge
  Base Construction and Web-scale Knowledge Extraction ({AKBC}-{WEKEX})}, pages
  95--100, Montr{\'e}al, Canada. Association for Computational Linguistics.

\bibitem[{Niu and Bansal(2018)}]{niu-bansal-2018-polite}
Tong Niu and Mohit Bansal. 2018.
\newblock \href {https://doi.org/10.1162/tacl_a_00027} {Polite dialogue
  generation without parallel data}.
\newblock \emph{Transactions of the Association for Computational Linguistics},
  6:373--389.

\bibitem[{Pennebaker et~al.(2015)Pennebaker, Boyd, Jordan, and
  Blackburn}]{pennebaker2015development}
James~W Pennebaker, Ryan~L Boyd, Kayla Jordan, and Kate Blackburn. 2015.
\newblock The development and psychometric properties of liwc2015.
\newblock Technical report, Austin, TX: University of Texas at Austin.

\bibitem[{Radford et~al.(2019)Radford, Wu, Child, Luan, Amodei, and
  Sutskever}]{radford2019language}
Alec Radford, Jeff Wu, Rewon Child, David Luan, Dario Amodei, and Ilya
  Sutskever. 2019.
\newblock Language models are unsupervised multitask learners.
\newblock \emph{OpenAI Blog}.

\bibitem[{Raffel et~al.(2019)Raffel, Shazeer, Roberts, Lee, Narang, Matena,
  Zhou, Li, and Liu}]{raffel2019exploring}
Colin Raffel, Noam Shazeer, Adam Roberts, Katherine Lee, Sharan Narang, Michael
  Matena, Yanqi Zhou, Wei Li, and Peter~J. Liu. 2019.
\newblock \href {http://arxiv.org/abs/1910.10683} {Exploring the limits of
  transfer learning with a unified text-to-text transformer}.

\bibitem[{Rush et~al.(2015)Rush, Chopra, and Weston}]{rush-etal-2015-neural}
Alexander~M. Rush, Sumit Chopra, and Jason Weston. 2015.
\newblock \href {https://doi.org/10.18653/v1/D15-1044} {A neural attention
  model for abstractive sentence summarization}.
\newblock In \emph{Proceedings of the 2015 Conference on Empirical Methods in
  Natural Language Processing}, pages 379--389, Lisbon, Portugal. Association
  for Computational Linguistics.

\bibitem[{Shang et~al.(2019)Shang, Li, Fu, Bing, Zhao, Shi, and
  Yan}]{shang-etal-2019-semi}
Mingyue Shang, Piji Li, Zhenxin Fu, Lidong Bing, Dongyan Zhao, Shuming Shi, and
  Rui Yan. 2019.
\newblock \href {https://doi.org/10.18653/v1/D19-1499} {Semi-supervised text
  style transfer: Cross projection in latent space}.
\newblock In \emph{Proceedings of the 2019 Conference on Empirical Methods in
  Natural Language Processing and the 9th International Joint Conference on
  Natural Language Processing (EMNLP-IJCNLP)}, pages 4937--4946, Hong Kong,
  China. Association for Computational Linguistics.

\bibitem[{Song et~al.(2019)Song, Tan, Qin, Lu, and Liu}]{song19d}
Kaitao Song, Xu~Tan, Tao Qin, Jianfeng Lu, and Tie-Yan Liu. 2019.
\newblock \href {http://proceedings.mlr.press/v97/song19d.html} {{MASS}: Masked
  sequence to sequence pre-training for language generation}.
\newblock In \emph{Proceedings of the 36th International Conference on Machine
  Learning}, volume~97 of \emph{Proceedings of Machine Learning Research},
  pages 5926--5936, Long Beach, California, USA. PMLR.

\bibitem[{Yan et~al.(2020)Yan, Qi, Gong, Liu, Duan, Chen, Zhang, and
  Zhou}]{yan2020prophetnet}
Yu~Yan, Weizhen Qi, Yeyun Gong, Dayiheng Liu, Nan Duan, Jiusheng Chen, Ruofei
  Zhang, and Ming Zhou. 2020.
\newblock \href {http://arxiv.org/abs/2001.04063} {Prophetnet: Predicting
  future n-gram for sequence-to-sequence pre-training}.

\bibitem[{Yang et~al.(2018)Yang, Hu, Dyer, Xing, and
  Berg-Kirkpatrick}]{NIPS2018_7959}
Zichao Yang, Zhiting Hu, Chris Dyer, Eric~P Xing, and Taylor Berg-Kirkpatrick.
  2018.
\newblock \href
  {http://papers.nips.cc/paper/7959-unsupervised-text-style-transfer-using-language-models-as-discriminators.pdf}
  {Unsupervised text style transfer using language models as discriminators}.
\newblock In S.~Bengio, H.~Wallach, H.~Larochelle, K.~Grauman, N.~Cesa-Bianchi,
  and R.~Garnett, editors, \emph{Advances in Neural Information Processing
  Systems 31}, pages 7287--7298. Curran Associates, Inc.

\bibitem[{Zhang et~al.(2019)Zhang, Zhao, Saleh, and Liu}]{zhang2019pegasus}
Jingqing Zhang, Yao Zhao, Mohammad Saleh, and Peter~J. Liu. 2019.
\newblock \href {http://arxiv.org/abs/1912.08777} {Pegasus: Pre-training with
  extracted gap-sentences for abstractive summarization}.

\bibitem[{Zhang* et~al.(2020)Zhang*, Kishore*, Wu*, Weinberger, and
  Artzi}]{Zhang*2020BERTScore:}
Tianyi Zhang*, Varsha Kishore*, Felix Wu*, Kilian~Q. Weinberger, and Yoav
  Artzi. 2020.
\newblock \href {https://openreview.net/forum?id=SkeHuCVFDr} {Bertscore:
  Evaluating text generation with bert}.
\newblock In \emph{International Conference on Learning Representations}.

\bibitem[{Zhang et~al.(2018)Zhang, Ding, and Soricut}]{zhang-etal-2018-shaped}
Ye~Zhang, Nan Ding, and Radu Soricut. 2018.
\newblock \href {https://doi.org/10.18653/v1/N18-1138} {{SHAPED}:
  Shared-private encoder-decoder for text style adaptation}.
\newblock In \emph{Proceedings of the 2018 Conference of the North {A}merican
  Chapter of the Association for Computational Linguistics: Human Language
  Technologies, Volume 1 (Long Papers)}, pages 1528--1538, New Orleans,
  Louisiana. Association for Computational Linguistics.

\end{thebibliography}
\bibliographystyle{acl_natbib}

\appendix

\section{Training and Decoding Settings}\label{append:training_and_decoding}

\paragraph{Training.}

We train our {\it simplicity style scorer} and {\it ideology style scorers} for $10$ epochs. The peak learning rate is $1 \times 10^{-5}$ with a batch size of $32$.

The \textit{class-conditional language models} for simplicity and ideology are trained with a peak learning rate of $5 \times 10^{-4}$ until the perplexity stops dropping on the validation set. We limit the number of tokens in each batch to $2,048$.

All \textit{word unit predictors} are trained with a peak learning rate of $1 \times 10^{-4}$ until the loss on the validation set no longer drops. We use a batch size of $32$ for training. 

\paragraph{Decoding.}

We use beam search for decoding. A beam size of $5$ is used for all models except for the decoder state adjustment having a beam size $1$ (greedy decoding) to maintain a reasonable running time. Repeated trigrams are disabled for generation in all experiments. As suggested by \citet{lewis-etal-2020-bart} and \citet{yan2020prophetnet}, length penalties are set to $2.0$ and $1.0$ for summarization and headline generation, respectively. 
The minimum and maximum lengths are set for decoding at $55$ and $140$ for summarization, $0$ and $75$ for headline generation.

\section{Statistics on SemEval and Allsides}\label{append:semeval_allsides}

\begin{table}[ht]
    \centering
    \small
    \begin{tabular}{lccc}
    \toprule
        \textbf{Split} & \textbf{Left} & \textbf{Neutral} & \textbf{Right} \\
        \midrule
        Training & 122,449 & 86,472 & 138,064 \\
        Validation & 10,000 & 10,000 & 10,000 \\
        Test & 10,000 & 10,000 & 10,000 \\
        \bottomrule
    \end{tabular}
    \vspace{-1mm}
    \caption{Ideology distribution for training, validation and test set splits of SemEval.}
    \label{tab:semeval_ideology}
    \vspace{-2mm}
\end{table}

Each article in the SemEval dataset is labeled with a stance: \textit{left}, \textit{leaning left}, \textit{neutral}, \textit{leaning right}, or \textit{right}. Here we combine left and leaning-left articles into one bucket, and similarly for right and leaning-right articles.
The ideology distribution for training, validation and test splits are in Table~\ref{tab:semeval_ideology}.

In our human evaluation of ideology-controlled headline generation, we use data collected from Allsides. The Allsides news clusters are curated by editors. The stance labels for different publishers are provided by Allsides, which are synthesized from blind surveys, editorial reviews, third-party analyses, independent reviews, and community feedback. We collect all the Allsides news clusters by April 26, 2020. After removing empty clusters, the total number of news clusters is $4{,}422$. Among them, $2{,}985$ clusters contain articles written in all three stances. For each article in the cluster, we keep the first paragraph and pair it with the headline. We remove the bylines in the first paragraphs. 

\section{Additional Results for Headline Generation}\label{append:ideology_giga}

In Table~\ref{tab:ideology_giga_auto_eval}, we show the results of ideology-controlled headline generation on SemEval with BART fine-tuned on Gigaword~\cite{napoles-etal-2012-annotated}. 
Our methods are still effective, especially by using decoder states adjustment with style scorers.

\begin{table}[h]
\small
    \centering
    \setlength{\tabcolsep}{4pt}
    \begin{tabular}{lccccc}
    \toprule
        \multirow{2}{*}{\textbf{Model}} & \multicolumn{2}{c}{\textbf{Left}} & \multicolumn{2}{c}{\textbf{Right}} \\
        \cmidrule(lr){2-3}
        \cmidrule(lr){4-5}
        & \textbf{Ideol.} & \textbf{BERT} & \textbf{Ideol.} & \textbf{BERT}  \\
        \midrule
        \textsc{BART}  & 21.77 & 88.81 & 20.72 & 88.81 \\
        \midrule
        \multicolumn{6}{l}{\rule{0pt}{2ex}\bf Ours w/ Decoder State Adjustment}\\
        \textsc{Ideol. Scorer} & \textbf{39.61} & 87.96 & \textbf{34.14} & 87.89 \\
        \textsc{Ideol. CC-LM} & 27.38 & 87.79 & 22.21 & 87.76 \\
        \multicolumn{6}{l}{\bf Ours w/ Word Unit Prediction}\\
        \textsc{WordU} & 22.98 & 88.35 & 21.09 & 88.40 \\
        \textsc{Dynamic WordU} & 22.84 & 88.32 & 21.08 & 88.47 \\
        \bottomrule
    \end{tabular}
    \vspace{-1mm}
    \caption{
    Ideological headline generation results with BART fine-tuned on the Gigaword dataset.  
    }
    \label{tab:ideology_giga_auto_eval}
    \vspace{-2mm}
\end{table}

\section{Human Evaluation Guidelines}

We include the evaluation guidelines for summarization and headline generation in Figures~\ref{tab:summ_human_eval} and~\ref{tab:headline_human_eval}. 

\begin{figure*}[t]
	\fontsize{10}{11}\selectfont
    \centering
    \begin{tabular}{lp{120mm}}
         \toprule
         \multicolumn{2}{c}{\textbf{Article}} \\
         \midrule
         & There was no special treatment for Lewis Ferguson at Paul Nicholls’s yard on Thursday morning. The 18-year-old was mucking out the stables as usual, just a cut on the nose to show for the fall which has made him an internet sensation. Ferguson’s spectacular double somersault fall from the favourite Merrion Square in the 4.20pm at Wincanton has been watched hundreds of thousands of times online. But he was back riding out and is undeterred from getting back in the saddle. Amateur jockey Lee Lewis Ferguson has just a cut on his nose to show for his ordeal . Teenager Ferguson was flung from his horse in spectacular fashion at Wincanton . ‘It was just a blur,’ he said. ‘I couldn’t work out what had happened until I got back to the weighing room and watched the replay. All the other jockeys asked me if I was all right and stuff, they all watched with me and looked away in horror.  (....) \\
         \midrule
         \multicolumn{2}{c}{\textbf{Informativeness:}} \\
         \midrule
         \rowcolor{lightgray!30}
          1 & Not relevant to the article \\
          & e.g., \textit{``Paul Nicholl's yard will start its expansion in December. The expansion plan was carried out six months ago."} \\
          \rowcolor{lightgray!30}
          3 & Relevant, but misses the main point \\
          & e.g., \textit{``Amateur jockey Lee Lewis Ferguson has just a cut on his nose to show for his ordeal . ‘It was just a blur,’ he said."} \\
          \rowcolor{lightgray!30}
          5 &  Successfully captures the main point and most of the important points. \\
          & e.g., \textit{``Lewis Ferguson was mucking out the stables as usual on Thursday. Favourite Merrion Square threw jockey in a freak fall on Wednesday."} \\
          \midrule
         \multicolumn{2}{c}{\textbf{Fluency:}} \\
         \midrule
         \rowcolor{lightgray!30}
         1 & Summary is full of garbage fragments and is hard to understand  \\
         & e.g., \textit{``18 year old nose. to cut show nose. the horse fashion, as to"} \\
         \rowcolor{lightgray!30}
         2 & Summary contains fragments, missing components but has some fluent segments \\
         & e.g., \textit{``Lewis Ferguson out on Thursday. threw jockey on Wednesday."} \\
         \rowcolor{lightgray!30}
         3 & Summary contains some grammar errors but is in general fluent\\
         & e.g., \textit{``Lewis Ferguson was muck out the stables as usual onThursday. The Merrion Square threw jockey jockey in a freak fall on Wednesday. His spectacular doublesomersault fall made him internetsensation."} \\
         \rowcolor{lightgray!30}
         4 & Summary has relatively minor grammatical errors \\
         & e.g., \textit{``Lewis Ferguson was mucking out the stables as usual on in Thursday. Favourite Merrion Square threw jockey ina freak fall on Wednesday. His spectacular double somersault fall made him internet sensation."} \\
         \rowcolor{lightgray!30}
         5 & Fluent Summary \\
         & e.g., \textit{"Lewis Ferguson was mucking out the stables as usual on Thursday. Favourite Merrion Square threw jockey in a freak fall on Wednesday. His spectacular double somersault fall made him internet sensation."} \\

         \midrule
         \multicolumn{2}{c}{\textbf{Simplicity:}} \\
         \midrule
         \rowcolor{lightgray!30}
         Bad & The summary uses complex words that can be replaced with simpler ones in almost all sentences and complex syntax structures (e.g., two or more clauses in a sentence) \\
         &  e.g., \textit{``Lewis Ferguson was thrown by Merrion Square and made a spectacular double somersault fall which gathered millions of views online, making him internet sensation. But he was back riding out and is undeterred from getting back in the saddle, just a cut on the nose to show for the fall ."} \\
         \rowcolor{lightgray!30}
         Moderate & The summary uses at most one complex words that can be replaced with simpler ones per sentence, and uses syntax structures with at most one clause in a sentence \\
         &  e.g., \textit{``Lewis Ferguson fell from Merrion Square. His spectacular double somersault fall made him internet sensation. But he was back riding out and is not afraid of getting back in the saddle."} \\
         \rowcolor{lightgray!30}
         Good & The summary almost always uses simple and common words and simple syntax structures (e.g., no clause or at most one clause in the whole summary) \\
         & e.g., \textit{"Lewis Ferguson fell from his horse on Wednesday. His eye-catching double flip fall made him famous on the Internet. He was back to the yard. He is not afraid of getting back in the saddle."} \\
         \bottomrule
    \end{tabular}
    \caption{Sample summaries with explanations on human evaluation aspect scales and examples of summaries at different simplicity levels.}
    \label{tab:summ_human_eval}
\end{figure*}

\begin{figure*}
	\fontsize{10}{11}\selectfont
    \centering
    \begin{tabular}{lp{120mm}}
         \toprule
         \multicolumn{2}{c}{\textbf{Paragraph}} \\
         \midrule
         & US President Donald Trump has said he is going to halt funding to the World Health Organization (WHO) because it has "failed in its basic duty" in its response to the coronavirus outbreak. \\
         \midrule
         \multicolumn{2}{c}{\textbf{Relevance:}} \\
         \midrule
         \rowcolor{lightgray!30}
          1 & The headline does not contain any information related to the input \\
          & e.g., \textit{``'a hateful act': what we know about the ft. lauderdale airport shooting"} \\
          \rowcolor{lightgray!30}
          2 & The headline contains some relevant event or person in the paragraph, but the topic is largely irrelevant \\
          & e.g., \textit{``trump: i don't take questions from cnn"} \\
          \rowcolor{lightgray!30}
          3 & The headline includes the main point of the paragraph, but have a different focus \\
          & e.g., \textit{``health experts condemn donald trump’s who funding freeze: ‘crime against humanity’"} \\
          \rowcolor{lightgray!30}
          4 & The headline captures the main point of the paragraph, but contains some information that cannot be inferred from the paragraph \\
          & e.g., \textit{``trump cuts off u.s. funding to who, pending review"} \\
          \rowcolor{lightgray!30}
          5 & The content of the headline and the paragraph are well aligned \\
          & e.g., \textit{``coronavirus: us to halt funding to who, says trump"} \\
         \bottomrule
    \end{tabular}
    \caption{Sample headlines with explanations on human evaluation aspect scales.}
    \label{tab:headline_human_eval}
\end{figure*}

\section{Sample Outputs}\label{append:sample_output}

Additional outputs are in Figures~\ref{fig:example_summ} and \ref{fig:example_hg}.

\begin{figure*}[!th]
    \small
    \setlength{\tabcolsep}{0pt}
    \centering
    \renewcommand{\arraystretch}{1.2}
    \begin{tabular}{p{0.98\textwidth}}
        \toprule
        \textbf{Example A} \\
        \midrule
        \textbf{Article:} $\pmb{\big [}$ No Tiger Woods and no Rory McIlroy in the field is proving no problem \textcolor{gray}{in Houston}, where record numbers have descended on the small suburb of Humble for the Shell Houston Open. $\pmb{\big ]}$  $\ldots$ It’s 18 months now since Spieth attracted criticism for giving up on university and joining the PGA Tour as the 809th ranked player in the world. Spieth, only 21, has since risen to number four in the world. Funnily enough, no-one’s criticising him anymore. Spectators have flocked to see Texas local boy Jordan Spieth in action at the Shell Houston Open. Spieth's fellow American Phil Mickelson has also \textcolor{cyan!10!blue}{\textbf{impressed}} in \textcolor{cyan!60!black}{\textbf{the run up to the Masters}} . $\ldots$ \\
        \midrule
        \textbf{Reference:} Record numbers have flocked to see local boy Jordan Spieth in action . Spieth and Phil Mickelson have been gearing up for the Masters . Texas-born Spieth quit university in order to concentrate on golf career . \\
        \textbf{Decoder State Adjustment:} $\pmb{\big [}$ Tiger Woods and Rory McIlroy are not in the field at the Shell Houston Open. Record numbers have descended on the small suburb of Humble. $\pmb{\big ]}$ Phil Mickelson and local boy Jordan Spieth have impressed in the run up to the Masters. Spieth, only 21, has risen to number four in the world. (\textit{reorganize sentences}) \\
        \textbf{Word Unit Prediction:} The Shell Houston Open is the last tournament before the Masters. Phil Mickelson and Jordan Spieth are both \textcolor{cyan!10!blue}{\textbf{playing well}} at \textcolor{cyan!60!black}{\textbf{the tournament}}. Spieth, only 21, is now number four in the world after 18 months on the PGA Tour. The American has finished in the top seven only twice this season. \\
        \midrule \textbf{Example B} \\
        \midrule
        \textbf{Article:} \ldots $\pmb{\big [}$Raikkonen's contract finishes at the end of the current Formula One season, \textcolor{gray}{although} there is an option for 2016 \textcolor{gray}{providing both parties are in agreement.}$\pmb{\big ]}$ The Finn stated this week he has never been happier working with a team in his entire F1 career, although his form to date has not matched that of team-mate Sebastian Vettel. $\pmb{\big [}$ Kimi Raikkonen has been \textcolor{cyan!10!blue}{\textbf{urged} to \textbf{improve his performances}} \textcolor{gray}{if he wants to stay at Ferrari}. $\pmb{\big ]}$ \ldots  \\
        \midrule
        \textbf{Reference:} Kimi Raikkonen's deal with Ferrari expires at the end of the season . Team principal Maurizio Arrivabene wants Finn to improve performances . Maranello driver has struggled to replicate team-mate Sebastian Vettel . \\
        \textbf{Decoder State Adjustment:} $\pmb{\big [}$ Kimi Raikkonen's contract with Ferrari ends at the end of the season. There is an option for him to stay with the Maranello marque in 2016. $\pmb{\big ]}$ Raikk Finnish has said he has never been happier working with a team. $\pmb{\big [}$ But team principal Maurizio Arrivabene has urged him to improve. $\pmb{\big ]}$ \\
        \textbf{Word Unit Prediction:} Kimi Raikkonen is out of contract with Ferrari at the end of the season. The Finn has said he has never been better at a team in his F1 career. But team-mate Sebastian Vettel has been in better form so far this season. Maurizio Arrivabene has said that \textcolor{cyan!10!blue}{\textbf{results will have to go up}} if he wants to be at Ferrari for another season.\\
        \midrule \textbf{Example C} \\
        \midrule
        \textbf{Article:} Three British citizens arrested in the United Arab Emirates after they were found plane spotting near Fujairah airport are to be released Monday, their lawyer, Nasser al-Hashem, tells CNN. The three have been in jail since February 22. \ldots No charges were filed, there will be no travel ban, and the men will not face \textcolor{cyan!10!blue}{\textbf{deportation}}, he said.  $\pmb{\big [}$\textcolor{gray}{Conrad Clitheroes, 54, and Gary Cooper, 45,}$\pmb{\big ]}$ were on a five-day visit to the UAE from Manchester when they were arrested. The third man, $\pmb{\big [}$\textcolor{gray}{Neil Munro}$\pmb{\big ]}$, is a British national who lives in the UAE. As a hobby, plane spotters view and photograph aircraft around the world. \\
        \midrule
        \textbf{Reference:} Three British men won't be charged or deported, their lawyer says. They were arrested after plane spotting near Fujairah airport and have been in jail since February 22. \\
        \textbf{Decoder State Adjustment:} Three Britons were arrested in the United Arab Emirates in February. They were found plane spotting near Fujairah airport. The men will be released Monday, their lawyer says. No charges were filed, and the men will not face deportation.. The three men were on a five-day visit to the UAE from Manchester. (\textit{omit the name}) \\
        \textbf{Word Unit Prediction:} The three have been in jail since February 22. They were arrested near Fujairah airport in the United Arab Emirates. The three will be released on Monday. No charges will be brought against them, and they will not \textcolor{cyan!10!blue}{\textbf{be sent back to}} the U.K. or other countries. \\
        \bottomrule
    \end{tabular}
    \caption{ 
    Examples of document summarization on CNN/DM dataset. We highlight lexical simplifications with blue colors. Texts removed for simplification by decoder state adjustment are in \textcolor{gray}{\textbf{gray}} or explained in \textit{italics}.
    }
    \label{fig:example_summ}
\end{figure*}

\begin{figure*}
    \small
    \setlength{\tabcolsep}{3pt}
    \centering
    \begin{tabular}{lp{0.40\textwidth}p{0.40\textwidth}}
    \toprule
    \textbf{Example A} \\
    \midrule
         & \multicolumn{2}{p{0.80\textwidth}}{\textbf{Paragraph:} Acting chief of staff Mick Mulvaney says President Trump willing to accept a barrier made of steel} \\
        \midrule
        \textbf{\textsc{Reference}} & \cellcolor{LeftBlue!60} mulvaney: saturday shutdown meeting `did not make much progress' & \cellcolor{red!15} mick mulvaney: trump willing to take concrete wall `off the table' \\
        \midrule
        \textbf{\textsc{Reranking}} & \cellcolor{LeftBlue!60} mick mulvaney says trump willing to accept a barrier made of steel & \cellcolor{red!15} mick mulvaney: trump willing to accept steel barrier \\
        \midrule
        \textbf{\textsc{LblCtrl}} & \cellcolor{LeftBlue!60} mick mulvaney: trump willing to accept barrier made of steel & \cellcolor{red!15} mick mulvaney: trump willing to accept barrier made of steel \\
        \midrule
        \textbf{\textsc{Ideol.Scorer}} & \cellcolor{LeftBlue!60} trump's budget proposal would increase the number of \textbf{military contractors} in the us & \cellcolor{red!15} trump wants to build a wall, \textbf{and he's willing to pay for it} \\
        \midrule
        \textbf{\textsc{Dynamic WordU}} & \cellcolor{LeftBlue!60} trump wants a border wall, \textbf{but it's not all about the wall}  & \cellcolor{red!15} mick mulvaney: trump willing to accept `steel' border wall \\
        \midrule
        \textbf{Example B} \\
        \midrule
        & \multicolumn{2}{p{0.80\textwidth}}{\textbf{Paragraph:} Rep. Paul Ryan accused President Barack Obama of emboldening Iran and those storming U.S. embassies abroad while curtailing individual freedoms at home, during a speech here to a gathering of religious conservatives.} \\
        \midrule
        \textbf{\textsc{Reference}} & \cellcolor{LeftBlue!60} paul ryan hits obama on national security: if we project weakness, they come & \cellcolor{red!15} ryan to values voters: ``american foreign policy needs moral clarity" \\
        \midrule
        \textbf{\textsc{Reranking}} & \cellcolor{LeftBlue!60} paul ryan accuses obama of emboldening iran, protesters & \cellcolor{red!15} paul ryan: obama emboldens iran \textbf{healthcare bill} \\
        \midrule
        \textbf{\textsc{LblCtrl}} & \cellcolor{LeftBlue!60} paul ryan: obama emboldening iran & \cellcolor{red!15} ryan: obama emboldened iran, embassy protesters \\
        \midrule
        \textbf{\textsc{Ideol.Scorer}} & \cellcolor{LeftBlue!60} paul ryan accuses obama of emboldening iran, protesters at \textbf{religious conservatives}' gathering & \cellcolor{red!15} ryan: obama emboldening iran, protesters while \textbf{curtailing freedoms at home} \\
        \midrule
        \textbf{\textsc{Dynamic WordU}} & \cellcolor{LeftBlue!60} paul ryan to \textbf{religious conservatives}: obama has ‘emboldened’ iran & \cellcolor{red!15} paul ryan: obama has ‘emboldened’ iran, protesters \\
        \midrule
        \textbf{Example C} \\
        \midrule
        & \multicolumn{2}{p{0.80\textwidth}}{\textbf{Paragraph:} The FBI on Wednesday issued an extraordinary public statement condemning the Republican push to release a classified memo that alleges surveillance abuses at the Department of Justice.} \\
        \midrule
        \textbf{\textsc{Reference}} & \cellcolor{LeftBlue!60} opinion: why trump is so eager to release the nunes memo & \cellcolor{red!15} trump to declassify infamous fisa memo \\
        \midrule
        \textbf{\textsc{Reranking}} & \cellcolor{LeftBlue!60} the fbi just responded to the \textbf{gop}’s push to release the memo & \cellcolor{red!15} fbi condemns \textbf{gop} push to release classified memo \\
        \midrule
        \textbf{\textsc{LblCtrl}} & \cellcolor{LeftBlue!60} the fbi just issued a public statement condemning the release of the \textbf{republican} memo & \cellcolor{red!15} fbi condemns \textbf{gop} push to release classified memo \\
        \midrule
        \textbf{\textsc{Ideol.Scorer}} & \cellcolor{LeftBlue!60} the fbi just released a statement condemning the release of the \textbf{republican} memo & \cellcolor{red!15} fbi releases statement condemning release of \textbf{russia} memo \\
        \midrule
        \textbf{\textsc{Dynamic WordU}} & \cellcolor{LeftBlue!60} fbi condemns \textbf{gop} push to release classified memo on \textbf{russia} & \cellcolor{red!15} fbi condemns \textbf{gop} push to release memo on \textbf{surveillance abuses} \\
        \bottomrule
    \end{tabular}
    \caption{Examples of ideology-controlled headline generation. Best viewed in color. The left panel (shaded in \textcolor{LeftBlue!90!blue}{\bf blue}) shows headlines generated with control toward the left stance. The right panel (\textcolor{red!60}{\bf red}) shows headlines generated with control toward the right. We highlight words that are commonly used with the corresponding stances in \textbf{bold}.}
    \label{fig:example_hg}
\end{figure*}

\end{document}